%% file: main.tex
\pdfoutput=1
\documentclass[letterpaper,journal]{IEEEtran}
\usepackage{amsmath,amsfonts,amssymb}
\usepackage{algorithmic}
\usepackage{algorithm}
\usepackage{array}
\usepackage[caption=false,font=normalsize,labelfont=sf,textfont=sf]{subfig}
\usepackage{textcomp}
\usepackage{stfloats}
\usepackage{url}
\usepackage{verbatim}
\usepackage{graphicx}
\usepackage{cite}
\usepackage{booktabs}
\usepackage{multirow}
\usepackage{xcolor}
\usepackage{xspace}
\graphicspath{{figures/}{../figures/}{../}}


\newcommand{\tablebody}{\footnotesize\renewcommand{\arraystretch}{0.96}}
\newcommand{\tabnote}[1]{%
\vspace{1pt}\par\begin{minipage}{0.98\linewidth}%
\footnotesize\emph{Note:} #1%
\end{minipage}}

\newcommand{\methodname}{RegisterBridgeMM\xspace}
\newcommand{\rwpr}{RWPR\xspace}
\newcommand{\rcrs}{RCRS\xspace}
\newcommand{\mac}{MAC\xspace}

\begin{document}

\title{RegisterBridgeMM: A Register-Centric Framework for RGB--Infrared Object Detection}

\author{\small
Zian Wang$^{1*}$, Hangchuan Liang$^{1*}$, Yuehua Chen$^{1*}$,
Changchun Li$^{1}$, Chaoyi Guo$^{2}$, Mingzhe Liu$^{3}$, Fangming~Gu$^{1\dagger}$\\
$^{1}$Jilin University, China \quad
$^{2}$Taiyuan University of Technology, China \quad
$^{3}$Shenzhen University, China\\
$^{*}$Equal contribution. \quad $^{\dagger}$Corresponding author.}

\markboth{}{}

\maketitle

\input{sections/abstract}
\input{sections/keywords}
\input{sections/introduction}
\input{sections/related_work}
\input{sections/method}
\input{sections/experiments}
\input{sections/conclusion}
\input{sections/references}

\end{document}

%% file: sections/abstract.tex
\begin{abstract}
RGB--infrared (RGB--IR) object detection benefits from complementary visible and thermal cues, but effective fusion remains challenging under illumination changes, weather variation, and cluttered scenes. Existing RGB--IR fusion methods often trade expressive patch-level interaction for lighter but more constrained adaptation mechanisms. We empirically observe that pretrained register tokens contain both modality-shared and modality-specific information on paired RGB--IR inputs, suggesting that they can serve as a compact substrate for cross-modal communication. Building on this observation, we propose \methodname, a register-mediated fusion framework organized as a three-stage register lifecycle. \emph{Aggregate} preserves per-modality register summarization inherited from pretraining; \emph{Bridge} performs bidirectional register-to-patch reading with consensus-residual regulation; and \emph{Project} translates the resulting register summary into spatially adaptive calibration of patch features. This register pathway avoids dense patch-to-patch cross-modal interaction while preserving the pretrained patch representation. With both backbone streams frozen, \methodname achieves the highest mAP$_{50\text{-}95}$ among the evaluated methods on all four benchmarks: LLVIP, M3FD, DroneVehicle, and FLIR-Aligned.
\end{abstract}

%% file: sections/keywords.tex

%% file: sections/introduction.tex
\section{Introduction}
\IEEEPARstart{M}{ultimodal} object detection that combines RGB and infrared (IR) imagery has become indispensable for all-weather perception in scenarios such as traffic monitoring, low-light surveillance, and autonomous driving. The two modalities are highly complementary: RGB provides rich texture and color cues but degrades sharply under low illumination, fog, or rain; IR captures thermal radiation and remains stable across illumination conditions, yet lacks fine-grained texture and category-discriminative detail~\cite{ref_llvip,ref_m3fd,ref_dronevehicle,ref_flir_adas}. Empirical evidence shows that single-modality detectors lose $10$--$20$ mAP at night, while properly fused models largely recover this gap~\cite{ref_cft,ref_icafusion,ref_cmpd}. Designing the cross-modal fusion mechanism is, therefore, a central question in this field.

The mainstream trajectory of RGB--IR fusion has largely focused on \emph{patch tokens} as the primary fusion substrate. From early CNN-based methods that exchange feature maps spatially~\cite{ref_gaff,ref_cmpd,ref_cmhf}, to recent ViT-based methods built on patch-wise cross-attention~\cite{ref_cft,ref_icafusion,ref_gmdetr,ref_fusionmamba,ref_wavemamba}, the underlying assumption has often been that cross-modal interaction should occur at the spatial granularity of patches. Two representative strategies have emerged under this assumption, each with practical limitations. The first, \emph{high-fidelity fusion}, computes dense $O(N^2)$ patch-to-patch cross-attention between the two streams, whose cost grows quadratically with the number of patches $N$ and becomes a dominant overhead at detection-scale resolutions. The second, \emph{lightweight injection}, reduces this cost by injecting features from one modality into the other in a master--slave configuration~\cite{ref_unirgbir,ref_slgnet}, but its asymmetric structure can be less robust when the auxiliary modality becomes more informative, such as IR at night. These limitations motivate a different design space in which cross-modal communication is routed through compact global tokens rather than dense patch-token interactions.

\begin{figure}[!t]
\centering
\subfloat[Per-token modality similarity: pretrained vs.\ trained.]{
\includegraphics[width=0.95\columnwidth]{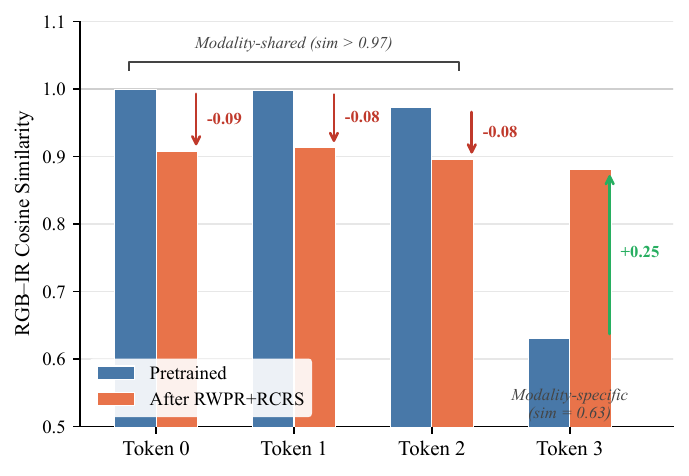}
\label{fig:intro_a}}\\[-1mm]
\subfloat[Cross-attention selectivity across layers.]{
\includegraphics[width=0.95\columnwidth]{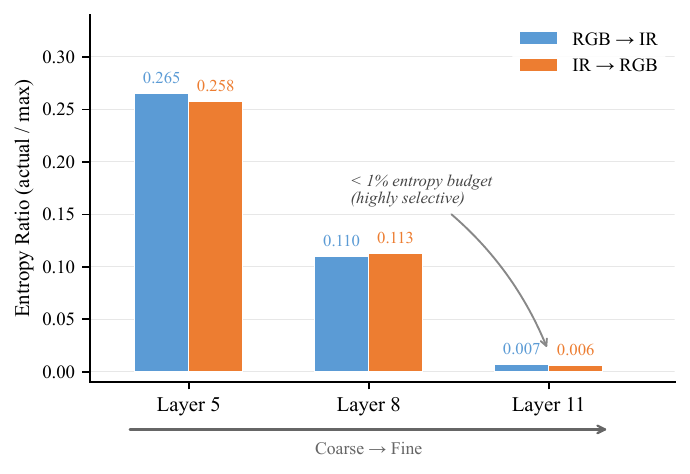}
\label{fig:intro_b}}
\caption{A pretrained register diagnostic and the learned bridge behavior of \methodname.
\textbf{(a)} Per-token cosine similarity between corresponding RGB and IR register embeddings of paired validation images. We compute the diagnostic by forwarding each RGB--IR pair through two independent DINOv3-B streams, extracting the four register positions, computing the RGB--IR cosine similarity at each corresponding register position, and averaging the values over the evaluated pairs. Without multimodal training (blue), three of the four register tokens align closely across modalities (similarity $>0.97$), while the fourth is more divergent (similarity $\approx 0.63$), revealing a $3{+}1$ modality-shared / modality-specific partition. After training with \rwpr$+$\rcrs under the same diagnostic protocol (orange), the highly aligned tokens become less collapsed ($-0.08$ similarity on average), while the divergent token moves closer to the regulated consensus ($+0.25$).
\textbf{(b)} Entropy ratio (actual / uniform-attention maximum) of a trained register-to-patch attention diagnostic at the feature readout layers. The ratio drops from $26\%$ at layer~5 to under $1\%$ at layer~11, showing coarse-to-fine selectivity; the two directions are nearly symmetric. Layer~11 therefore provides evidence of selective deep readout, whereas bridge injection is placed earlier so the updated registers can propagate through subsequent self-attention.}
\label{fig:intro}
\end{figure}

DINOv2~\cite{ref_dinov2} and DINOv3~\cite{ref_dinov3} introduce a small set of \emph{register tokens} alongside patch tokens, originally as a remedy for high-norm artifacts in ViT self-attention~\cite{ref_registers}. Throughout large-scale pretraining, registers absorb global context from patch tokens at every layer, ultimately functioning as compact summaries of the input image. Their potential role in RGB--IR fusion, however, remains underexplored. We make a related empirical observation: when paired RGB and IR images of the same scene are fed through two independent DINOv3 streams \emph{without multimodal training}, the four register tokens do not behave uniformly across modalities. Three register positions produce closely aligned embeddings (cosine similarity $>0.97$), while the fourth produces a more divergent embedding (cosine $\approx 0.63$). This pattern indicates a $3{+}1$ modality-shared / modality-specific organization (Fig.~\ref{fig:intro}(a), blue bars), suggesting a pretrained representation geometry that is compatible with consensus-residual fusion.

We further diagnose the trained register-to-patch bridge at the multi-scale feature readout layers. When registers query the patch tokens of the opposite modality, the attention entropy ratio drops from $26\%$ at layer~5 to under $1\%$ at layer~11 (Fig.~\ref{fig:intro}(b)), showing coarse-to-fine selectivity that is symmetric in both directions (RGB$\rightarrow$IR and IR$\rightarrow$RGB). This layer-11 behavior indicates that deeper registers provide highly selective cross-modal readout; it does not imply that the final layer should be used as a bridge-injection point. Because an updated register must pass through subsequent self-attention blocks to influence patch features, \methodname injects bridges before the deepest readout layer and later verifies this choice by layer-placement ablation.

Motivated by these observations, we propose \methodname, a register-centric framework for RGB--IR object detection that organizes register tokens into a three-stage register lifecycle. In the Aggregate stage, the frozen DINOv3 streams preserve per-modality register summarization inherited from pretraining. In the Bridge stage, \rwpr (Register-Write Patch-Read) lets registers read opposite-modality patches, while \rcrs (Register-Consensus / Residual-Split) decomposes the bridged registers into consensus and modality-specific residual components. In the Project stage, \mac (Modality-Aware Calibration) converts the register summary into spatially adaptive affine parameters that calibrate each modality's pyramid features before fusion.

Concretely, bridges are inserted at shallow-to-intermediate layers ($\{2,5,8\}$) rather than at the final readout layer, allowing updated registers to pass through subsequent frozen self-attention blocks before detection features are extracted. Throughout the backbone, cross-modal attention is routed through register queries only: patch tokens provide read-only keys and values and are never directly updated by cross-modal attention. With $k{=}4$ registers, the full three-layer bidirectional bridge uses a small fraction of the pairwise terms of one dense patch-wise cross-attention layer.

The contributions of this paper are summarized as follows.
\begin{itemize}
\item \textbf{A pretrained-register observation for RGB--IR fusion.} We identify a modality-shared and modality-specific organization in pretrained DINOv3 register tokens on paired RGB--IR inputs before multimodal training (Fig.~\ref{fig:intro}(a)), motivating registers as a structured communication substrate rather than generic latent queries.
\item \textbf{A register-mediated fusion framework.} Building on this observation, we propose \methodname, instantiating a three-stage register lifecycle---\emph{Aggregate}, \emph{Bridge}, and \emph{Project}---through \rwpr, \rcrs, and \mac. \rwpr performs efficient bidirectional register-to-patch reading, while \rcrs explicitly regulates the consensus-residual trade-off.
\item \textbf{Strong empirical performance with parameter-efficient adaptation.} With both DINOv3 streams frozen and only a lightweight set of trainable bridge modules, \methodname achieves competitive or superior performance compared with the evaluated methods on all four RGB--IR benchmarks: LLVIP~\cite{ref_llvip}, M3FD~\cite{ref_m3fd}, DroneVehicle~\cite{ref_dronevehicle}, and FLIR-Aligned~\cite{ref_flir_adas}.
\end{itemize}

%% file: sections/related_work.tex
\section{Related Work}
\label{sec:related}

\subsection{RGB--IR Object Detection}
RGB--IR object detection exploits the complementary sensing properties of visible and infrared imagery. RGB images provide texture, color, and category-discriminative appearance, whereas infrared images remain informative under weak illumination and adverse weather. Early multispectral detectors usually fused the two modalities at the image or feature-map level with convolutional backbones and hand-designed fusion blocks~\cite{ref_gaff}. Subsequent work refines this local fusion paradigm along several axes, including modality reliability estimation~\cite{ref_cmpd,ref_cmhf}, modality-imbalance mitigation~\cite{ref_mbnet}, and target- or alignment-aware feature aggregation~\cite{ref_tfdet,ref_yoloadaptor,ref_cpfm,ref_cofnet}. These methods established the value of cross-modal complementarity, but their local fusion operations provide limited control over long-range dependencies and modality-dependent reliability.

Recent work has moved from local convolutional fusion toward attention-, token-, and sequence-based interaction. Transformer-style detectors exchange or calibrate visible and infrared features through cross-attention modules~\cite{ref_cft,ref_icafusion,ref_irdfusion,ref_c2former,ref_gmdetr}, while broader multimodal ViT methods route or replace patch tokens across modalities~\cite{ref_tokenfusion}. Sequence-modeling approaches such as Fusion-Mamba and WaveMamba use state-space operators to capture long-range dependencies with improved efficiency~\cite{ref_fusionmamba,ref_wavemamba}. Benchmarking studies also show that training recipes and evaluation protocols can strongly influence multispectral detection comparisons~\cite{ref_msod_tricks}. Despite these advances, the primary fusion substrate remains feature maps or patch tokens. Dense patch-level interaction can be expressive, but its cost grows quickly with resolution; lighter designs often reduce the interaction to asymmetric injection or local feature modulation.

\methodname differs from these fusion paradigms by moving the primary cross-modal communication pathway from dense patch tokens to pretrained register tokens. Rather than applying cross-stream attention directly to all patch locations, \methodname lets a small set of register tokens query opposite-modality patches and then projects the resulting cross-modal summary back to patch features through calibration. This design preserves the benefits of patch-level evidence while routing communication through a compact register bottleneck.

\subsection{Frozen Vision Foundation Models and Adapter-Style Adaptation}
Large pretrained vision models provide strong representations for downstream dense prediction, but full fine-tuning can be parameter-intensive and may overwrite useful pretrained structure. Adapter-style adaptation addresses this by freezing most of the backbone and inserting lightweight trainable modules, as exemplified by parameter-efficient low-rank adaptation~\cite{ref_lora} and ViT-Adapter for dense prediction~\cite{ref_vitadapter}. In RGB--IR perception, this direction is particularly attractive because paired multispectral datasets are smaller and more domain-specific than the data used to train modern visual foundation models. UniRGB-IR demonstrates frozen-backbone adapter tuning for visible--infrared semantic tasks~\cite{ref_unirgbir}, and SLGNet introduces structural priors and language-guided modulation for multimodal detection~\cite{ref_slgnet}. These methods adapt strong pretrained representations with limited trainable capacity, but they still add cross-modal interaction outside the internal token organization of the pretrained backbone.

This leaves open a structural question: whether the pretrained model already contains a compact pathway that can be reused for cross-modal communication. \methodname answers this question by freezing the DINOv3 streams and learning only parameter-efficient modules around their register tokens. The goal is not only a small trainable parameter count, but also preservation of the pretrained patch representation and the register specialization observed before multimodal training.

\subsection{Register Tokens in Vision Transformers}
Vision Transformers represent images as patch-token sequences~\cite{ref_vit}. DINOv2 and DINOv3 extend this formulation with register tokens, a small set of additional tokens that absorb global information during self-attention~\cite{ref_dinov2,ref_dinov3}. Darcet \emph{et al.} showed that such registers help remove high-norm artifacts and improve the behavior of self-supervised ViT features~\cite{ref_registers}. In their original use, registers are primarily viewed as a stabilizing mechanism for visual representation learning: they collect global context and prevent patch tokens from carrying spurious artifacts.

Subsequent studies have begun to explore registers beyond artifact suppression, for example as features for robust adaptation, as test-time tokens that mimic trained registers, or as compact summaries for efficient vision--language models~\cite{ref_register_adapt,ref_untrained_registers,ref_compact_registers}. The downstream role of register tokens in multimodal detection, however, remains much less explored. Existing RGB--IR fusion methods generally treat pretrained ViT backbones as providers of patch features, while cross-modal interaction is performed by additional fusion blocks. Under this view, registers are either ignored or treated as implementation details of the backbone. Our empirical observation suggests a different interpretation. Before multimodal training, paired RGB and IR inputs passed through two independent frozen DINOv3 streams reveal a modality-shared and modality-specific organization in the register embeddings. After training the proposed bridge, the register-query attention over opposite-modality patches becomes increasingly selective across depth. These two findings play different roles: the former motivates registers as a pretrained communication substrate, whereas the latter analyzes the behavior learned by RWPR.

\methodname builds directly on this interpretation. RWPR uses registers as bidirectional cross-modal queries, RCRS makes their consensus-residual structure explicit, and MAC transfers the resulting register-level summary back to patch features before detection. In this way, the proposed framework turns registers from a passive component of pretrained ViTs into the organizing substrate of RGB--IR fusion.

%% file: sections/method.tex
\section{Method}
\label{sec:method}

\subsection{Overview}
\label{sec:method_overview}
The forward pipeline of \methodname follows a dual-stream RGB--IR design organized around register-mediated communication. Given a pair of spatially-aligned RGB and infrared (IR) inputs $\mathbf{I}_{R}, \mathbf{I}_{T} \in \mathbb{R}^{3 \times H \times W}$, the two modalities are fed into a dual-stream backbone instantiated from a pretrained DINOv3 ViT~\cite{ref_dinov3}. The two streams are initialized from the same pretrained weights and run independently, with all backbone parameters frozen throughout training. Following the DINOv3 input format, each stream represents an image as a token sequence consisting of one $\mathtt{[CLS]}$ token, $k{=}4$ \emph{register} tokens, and $N$ patch tokens. Throughout the paper we use $\mathbf{r}_{m,\ell} \in \mathbb{R}^{k \times D}$ and $\mathbf{p}_{m,\ell} \in \mathbb{R}^{N \times D}$ to denote, respectively, the register and patch tokens of modality $m{\in}\{R,T\}$ at layer $\ell$, with $D$ the embedding dimension.

\begin{figure*}[t]
\centering
\includegraphics[width=0.98\textwidth]{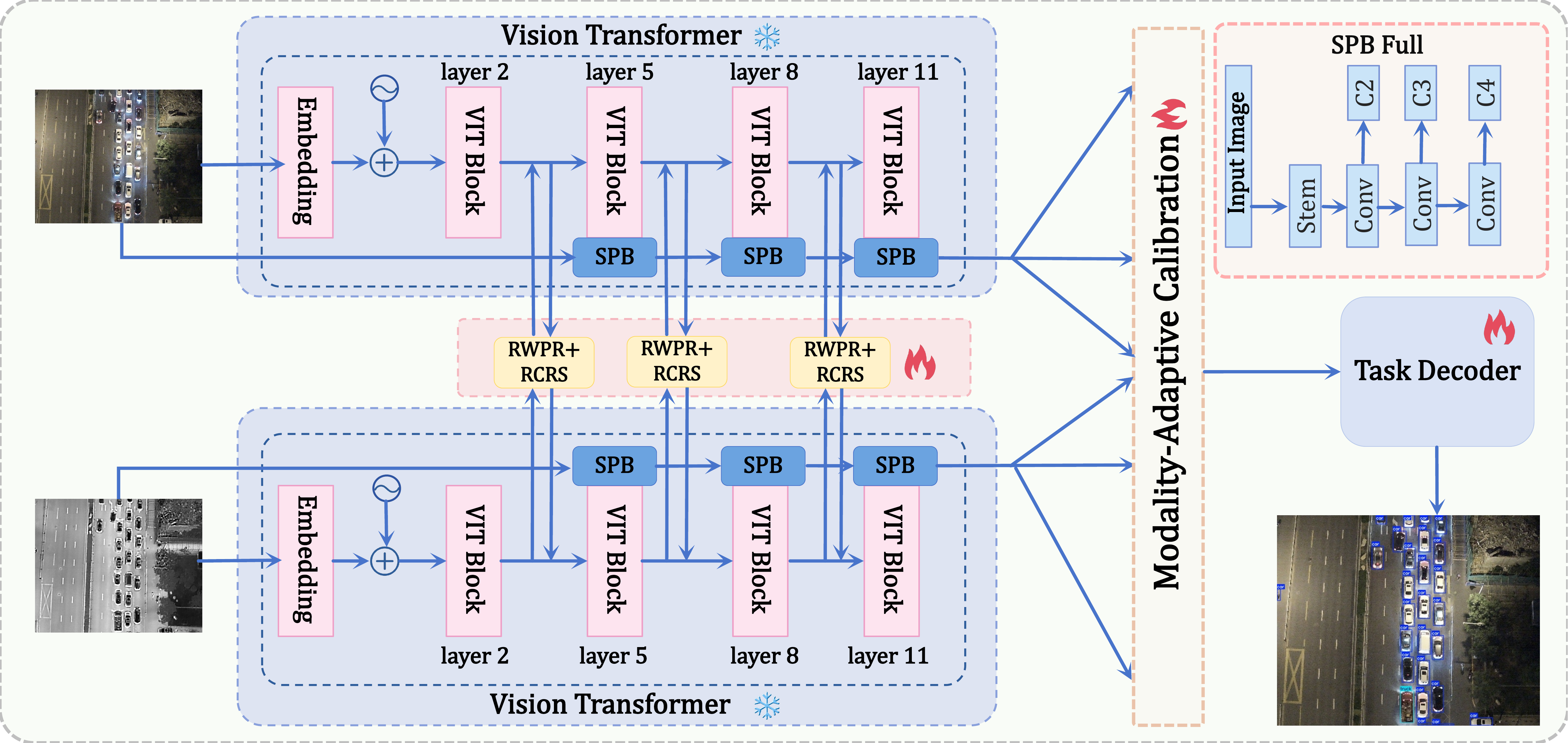}
\caption{Overview of \methodname. The three-stage register lifecycle: \textbf{Aggregate} (frozen DINOv3 streams accumulate per-modality register summaries), \textbf{Bridge} (RWPR performs bidirectional register-to-patch cross-attention; RCRS decomposes the result into consensus and modality-specific residuals), and \textbf{Project} (MAC translates the register-level CR summary into spatially adaptive affine calibration of the patch feature pyramids, followed by multi-scale fusion and the RT-DETR detection head). Dashed lines indicate frozen parameters.}
\label{fig:method_overview}
\end{figure*}

\methodname organizes the role of register tokens throughout the network as a three-stage \emph{register lifecycle}. Section~\ref{sec:method_aggregate} describes \textbf{Aggregate}, the inherited unimodal register summarization in the frozen streams. Section~\ref{sec:method_bridge} introduces \textbf{Bridge}, where \rwpr and \rcrs perform bidirectional register-to-patch reading and consensus-residual regulation. Section~\ref{sec:method_project} presents \textbf{Project}, where \mac converts the register summary into spatially adaptive feature calibration. An overview of the full architecture is shown in Fig.~\ref{fig:method_overview}.

\subsection{Stage I: Aggregate}
\label{sec:method_aggregate}
The Aggregate stage spans the early layers of each frozen DINOv3 stream and introduces no new module. At every encoder layer $\ell$, the $k$ register tokens of a stream jointly participate in self-attention with the $N$ patch tokens of the same modality. Because registers carry no spatial location and were trained at scale to absorb high-norm artifacts that would otherwise affect patch tokens~\cite{ref_registers}, they act as compact summarizers: from a sequence of $N$ spatially anchored patches, they distill a $k$-slot global digest of the modality. Formally, the per-layer update of a register reads
\begin{equation}
\mathbf{r}_{m,\ell+1} = \mathrm{SA}_{\ell}\big([\mathbf{c}_{m,\ell};\, \mathbf{r}_{m,\ell};\, \mathbf{p}_{m,\ell}]\big)\big|_{\text{reg}},
\label{eq:aggregate}
\end{equation}
where $\mathrm{SA}_{\ell}$ is the frozen self-attention block of layer $\ell$, $\mathbf{c}_{m,\ell}$ is the $\mathtt{[CLS]}$ token, and $|_{\text{reg}}$ extracts the $k$ register positions from the output sequence. After the first $L_0$ layers (with $L_0{=}2$ in our configuration, the index of the first injection layer), each stream's registers already encode a unimodal global summary that is independent of the other stream.

The motivation for keeping this stage parameter-free is structural rather than only economical. As shown in Fig.~\ref{fig:intro}(a), the registers produced by the pretrained Aggregate process already carry a latent modality-shared / modality-specific partition. A learnable intervention at this stage could weaken this partition before cross-modal bridging is introduced. We therefore freeze it and use the output $\{\mathbf{r}_{R,L_0},\mathbf{r}_{T,L_0}\}$ as the starting representation of the cross-modal pathway.

\subsection{Stage II: Bridge}
\label{sec:method_bridge}
After Aggregate, the two streams' registers each summarize their own modality, but they have never communicated. The Bridge stage addresses this at three shallow-to-intermediate layers $\mathcal{S}{=}\{2,5,8\}$, which are decoupled from the multi-scale feature layers $\{5,8,11\}$ later read out for the detection head. At each injection layer, \rwpr reads opposite-modality patches into registers, and \rcrs splits the bridged registers into consensus and modality residuals. The three bridges share architecture but not parameters. Although the layer-11 readout shows strong selectivity in Fig.~\ref{fig:intro}(b), we do not use the final readout layer as an injection point: a register update there has no subsequent self-attention block through which to influence patch features. The layer-placement ablation in Sec.~\ref{sec:exp} supports this separation between deep readout and earlier injection.

\noindent\textbf{Register-Write Patch-Read (RWPR).}
To exchange information without dense patch-to-patch attention, \rwpr uses the asymmetry between registers and patches. Patch-to-patch cross-attention costs $O(N^2)$ and becomes expensive at detection-scale resolutions, whereas register-to-register attention is cheap ($O(k^2)$) but has limited access to spatial detail. We therefore let the $k$ registers of one modality query the $N$ patches of the other, retaining dense spatial evidence at $O(kN)$ cost while routing the exchange through a compact register bottleneck.

At each injection layer $\ell{\in}\mathcal{S}$, we run two single-head cross-attention operators with independent parameters, denoted as $\mathrm{CA}$, one per direction:
\begin{align}
\boldsymbol{\Delta}_{R} &= \mathrm{CA}_{R{\rightarrow}T}^{(\ell)}\big(\mathbf{r}_{R,\ell},\,\mathbf{p}_{T,\ell},\,\mathbf{p}_{T,\ell}\big), \label{eq:rwpr_r}\\
\boldsymbol{\Delta}_{T} &= \mathrm{CA}_{T{\rightarrow}R}^{(\ell)}\big(\mathbf{r}_{T,\ell},\,\mathbf{p}_{R,\ell},\,\mathbf{p}_{R,\ell}\big), \label{eq:rwpr_t}
\end{align}
in which the registers of modality $m$ act as queries against the patch tokens of the opposite modality (acting as both keys and values). The two raw deltas are gated by a sigmoid-bounded residual coefficient $\alpha_{\ell}\,{=}\,\alpha_{\max}\sigma(\tilde{\alpha}_{\ell})$ and applied with LayerNorm:
\begin{equation}
\tilde{\mathbf{r}}_{m,\ell} = \mathrm{LN}\big(\mathbf{r}_{m,\ell} + \alpha_{\ell}\,\boldsymbol{\Delta}_{m}\big),\quad m\in\{R,T\},
\label{eq:rwpr_apply}
\end{equation}
with $\alpha_{\max}{=}1$ and $\sigma(\cdot)$ the logistic function. The logit $\tilde{\alpha}_{\ell}$ is initialized so that $\alpha_{\ell}{=}0.1$ at the start of training, which makes the bridge a near-identity transformation at initialization and avoids corrupting the pretrained representation in the early epochs.

Three properties distinguish RWPR from prior cross-modal interactions. \emph{(i) Complexity.} Each direction costs $O(kND)$, a fraction $k/N$ of dense patch-wise attention, which is small because $k{=}4$ while $N$ is in the thousands. \emph{(ii) Depth-dependent readout.} The diagnostic in Fig.~\ref{fig:intro}(b) shows that register-to-patch attention becomes increasingly selective at deeper feature readout layers. RWPR therefore injects information before the final readout stage, allowing updated registers to propagate through later self-attention while still benefiting from selective deep features. \emph{(iii) Backbone integrity.} RWPR updates only registers; patch tokens remain read-only keys and values, preserving the pretrained patch feature manifold while still allowing indirect cross-modal influence.

\noindent\textbf{Register-Consensus / Residual-Split (RCRS).}
RWPR augments each register with information from the opposite modality, yet a failure mode remains. As the two streams' registers absorb information from each other, they may drift toward an uncontrolled consensus, in which case the cross-stream channel carries similar global summaries and the modality-specific signal is diluted. RCRS is designed to make this trade-off explicit rather than to amplify modality-specific residuals. We therefore decompose the bridged register set into a consensus component shared by both modalities and a modality-specific residual, and re-mix the two with a learnable balance.

For each injection layer, after RWPR we form the per-layer consensus
\begin{equation}
\mathbf{c}_{\ell} = \tfrac{1}{2}\big(\tilde{\mathbf{r}}_{R,\ell} + \tilde{\mathbf{r}}_{T,\ell}\big),
\label{eq:rcrs_c}
\end{equation}
and re-parameterize each modality's registers as a consensus plus a scaled residual:
\begin{equation}
\mathbf{r}'_{m,\ell} = \mathbf{c}_{\ell} + \beta_{\ell}\big(\tilde{\mathbf{r}}_{m,\ell} - \mathbf{c}_{\ell}\big),\quad m\in\{R,T\},
\label{eq:rcrs_apply}
\end{equation}
where $\beta_{\ell}{=}\sigma(\tilde{\beta}_{\ell}) \in (0,1)$ is a scalar gate initialized at $\beta_{\ell}{=}0.5$. The two limits expose the meaning of the gate: $\beta_{\ell}{\to}0$ collapses both modalities to the consensus and erases all modality-specific content, whereas $\beta_{\ell}{\to}1$ preserves the original bridged registers without consensus regularization. The calibrated registers $\mathbf{r}'_{m,\ell}$ replace $\mathbf{r}_{m,\ell}$ in the token sequence and continue through the remaining frozen self-attention layers, so the effect of RCRS propagates through the entire upper backbone. The consensus vector $\mathbf{c}_{\ell}$ from the deepest injection layer is retained for use in Stage III.

RCRS turns the implicit shared-vs-specific structure of pretrained registers (Fig.~\ref{fig:intro}(a)) into an \emph{explicit} algebraic decomposition that the network can regulate. By exposing $\beta_{\ell}$ as a learnable scalar, RCRS lets the model choose the consensus--residual trade-off per layer rather than being committed to a hand-set ratio. After training, this decomposition reshapes the pretrained $3{+}1$ pattern into a task-adapted bridge, as further analyzed in Sec.~\ref{sec:exp}.

\subsection{Stage III: Project}
\label{sec:method_project}
After Stage II, the registers carry a structured cross-modal summary, but the patch features at the backbone outputs have not yet received it. Stage III addresses this through Modality-Aware Calibration (\mac), which converts the register summary into spatially adaptive affine parameters, followed by a light multi-scale fusion head for the RT-DETR decoder. Patch-only fusion methods often couple two tasks: \emph{estimating} how modalities complement each other at the scene level and \emph{applying} this estimate at each spatial location. Stages I and II produce a register-level summary for the first task; Stage III translates it into per-position patch calibration. $\ell^{\star}$ denote the deepest injection layer, $\mathbf{r}^{\star}_{m}{=}\mathbf{r}'_{m,\ell^{\star}}$ the post-RCRS registers used by \mac, and $\mathbf{c}^{\star}{=}\mathbf{c}_{\ell^{\star}}$ the corresponding consensus. We form a single CR summary vector by concatenating the consensus mean with the absolute magnitudes of the two modality residuals:
\begin{equation}
\mathbf{s}_{\mathrm{CR}} = \big[\,\overline{\mathbf{c}^{\star}};\;\overline{|\mathbf{r}^{\star}_{R}-\mathbf{c}^{\star}|};\;\overline{|\mathbf{r}^{\star}_{T}-\mathbf{c}^{\star}|}\,\big] \in \mathbb{R}^{3D},
\label{eq:cr_summary}
\end{equation}
where $\overline{(\cdot)}$ denotes the mean across the $k$ register positions. The first term encodes \emph{what the two modalities agree on}; the latter two encode \emph{how strongly each modality deviates from that agreement} at the register level. Spatial localization is supplied by the inter-modal gap proxy below.

Let $\mathbf{F}^{s}_{R}, \mathbf{F}^{s}_{T} \in \mathbb{R}^{C_s \times H_s \times W_s}$ denote the projected patch feature maps of the two modalities at pyramid level $s$. The modality-specific projection heads map both streams to the same channel width, and \mac first normalizes the two maps with per-stream GroupNorm, $\hat{\mathbf{F}}^{s}_{m} = \mathrm{GN}^{s}_{m}(\mathbf{F}^{s}_{m})$. It then produces a global affine seed from the CR summary through a shared two-layer MLP $\Phi(\cdot)$ followed by a per-scale linear projection $\mathbf{W}_s$:
\begin{equation}
\mathbf{seed}^{s} = \mathbf{W}_{s}\,\Phi(\mathbf{s}_{\mathrm{CR}}) \in \mathbb{R}^{C_s},
\label{eq:mac_seed}
\end{equation}
broadcast over space. To inject a spatial signal of \emph{where} the normalized modality features differ, we compute a learned inter-modal gap proxy and lift it through a $1{\times}1$ convolution:
\begin{equation}
\mathbf{gap}^{s} = \mathrm{Conv}_{1{\times}1}^{s}\big(\hat{\mathbf{F}}^{s}_{R} - \hat{\mathbf{F}}^{s}_{T}\big) \in \mathbb{R}^{C_s \times H_s \times W_s}.
\label{eq:mac_gap}
\end{equation}
The global seed and the spatial gap are summed to form a per-position conditioning field $\mathbf{field}^{s} = \mathbf{seed}^{s} + \mathbf{gap}^{s}$, which is fed into a per-scale affine head $\mathcal{A}^{s}$---a $1{\times}1$ convolution with zero-initialized weights---that outputs $4C_s$ channels chunked into the per-modality SPADE-style parameters $(\boldsymbol{\gamma}^{s}_{m}, \boldsymbol{\beta}^{s}_{m})$. The calibrated features are obtained as a bounded residual update,
\begin{equation}
\tilde{\mathbf{F}}^{s}_{m} = \mathbf{F}^{s}_{m} + \alpha_{s}\big(\boldsymbol{\gamma}^{s}_{m}\odot\hat{\mathbf{F}}^{s}_{m} + \boldsymbol{\beta}^{s}_{m}\big),\quad m\in\{R,T\},
\label{eq:mac_apply}
\end{equation}
where $\alpha_{s}{=}\sigma(\tilde{\alpha}_{s})$ is a per-scale sigmoid gate (initialized so that $\alpha_{s}{=}0.1$) and $\odot$ is element-wise multiplication. Because the affine head is zero-initialized and the gate is small at start, \mac is an exact identity at initialization and only departs from it as training proceeds.

The calibrated pyramids are then fused level-by-level. At each level $s$, the two modalities are concatenated and projected to the unified width through a $1{\times}1$ convolution $\mathcal{F}^{s}_{\mathrm{cat}}$, followed by a top-down residual that propagates information from coarser to finer levels:
\begin{equation}
\mathbf{F}^{s}_{\mathrm{fused}} = \mathcal{F}^{s}_{\mathrm{cat}}\big([\tilde{\mathbf{F}}^{s}_{R};\,\tilde{\mathbf{F}}^{s}_{T}]\big) + \lambda_{s}\,\mathcal{U}\big(\mathbf{F}^{s-1}_{\mathrm{fused}}\big),
\label{eq:mac_fuse}
\end{equation}
where $\mathcal{U}(\cdot)$ is bilinear upsampling to the resolution of level $s$ and $\lambda_{s}{=}\sigma(\tilde{\lambda}_{s})$ is a sigmoid-bounded scalar gate (the term is omitted for the coarsest level). The fused three-level pyramid is then consumed by an RT-DETR detection head~\cite{ref_rtdetr}.

\mac differs from channel-only FiLM-style modulations~\cite{ref_film} and feature statistics-only calibration because its affine field is both register-conditioned and spatially varying. The global seed transmits the cross-modal summary from Stages I--II, while the learned gap proxy supplies location-specific discrepancy cues. The zero-initialized head and bounded residual gate make \mac an identity mapping at initialization, which is important when both backbones are frozen.

%% file: sections/experiments.tex
\section{Experiments}
\label{sec:exp}

\begin{table*}[!t]
\centering
\caption{Comparison on M3FD.}
\label{tab:m3fd}
\tablebody
\begin{tabular*}{\textwidth}{@{\extracolsep{\fill}}llcccccccc@{}}
\toprule
Method & Venue & \multicolumn{2}{c}{Overall} & \multicolumn{6}{c@{}}{AP$_{50}$ by category} \\
\cmidrule(lr){3-4}\cmidrule(l){5-10}
 & & mAP$_{50}$ & mAP$_{50\text{-}95}$ & People & Car & Bus & Moto & Lamp & Truck \\
\midrule
TarDAL~\cite{ref_m3fd} & CVPR'22 & 80.5 & 54.1 & 81.5 & 94.8 & 81.3 & 69.3 & 87.1 & 68.7 \\
CDDFuse~\cite{ref_cddfuse} & CVPR'23 & 81.1 & 54.3 & 81.6 & 92.5 & 82.6 & 71.6 & 86.9 & 71.5 \\
IGNet~\cite{ref_ignet} & MM'23 & 81.5 & 54.5 & 81.6 & 92.8 & 82.4 & 73.0 & 86.9 & 72.1 \\
EMMA~\cite{ref_emma} & CVPR'24 & 82.9 & 55.4 & 82.0 & 93.5 & 83.2 & 77.7 & 87.7 & 73.5 \\
SuperFusion~\cite{ref_superfusion} & JAS'22 & 83.5 & 56.0 & 83.7 & 91.0 & 93.2 & 77.4 & 70.0 & 85.8 \\
KCDNet~\cite{ref_kcdnet} & TIM'24 & 83.2 & 56.3 & 83.3 & 90.9 & 88.4 & 84.1 & 80.3 & 72.1 \\
ICAFusion~\cite{ref_icafusion} & PR'24 & 87.1 & 57.0 & 83.3 & 93.4 & 93.4 & 75.0 & 87.6 & \textbf{90.0} \\
MMFN~\cite{ref_mmfn} & TITS'23 & 86.2 & 57.4 & 83.0 & 93.2 & 92.1 & 73.7 & 87.6 & 87.4 \\
Fusion-Mamba~\cite{ref_fusionmamba} & TMM'25 & 88.0 & 61.9 & 84.3 & 92.9 & 94.2 & 80.5 & 87.5 & 88.8 \\
LCAFNet~\cite{ref_lcafnet} & PR'26 & 91.1 & 59.3 & 90.2 & 94.3 & 92.4 & 86.0 & \textbf{94.2} & \underline{89.5} \\
WaveMamba~\cite{ref_wavemamba} & ICCV'25 & \underline{92.1} & \underline{64.4} & \underline{92.6} & \underline{95.9} & \underline{95.6} & \underline{86.1} & 93.8 & 88.6 \\
\midrule
\textbf{\methodname} & -- & \textbf{92.6} & \textbf{64.9} & \textbf{92.9} & \textbf{96.1} & \textbf{95.8} & \textbf{86.5} & \underline{94.0} & 89.1 \\
\bottomrule
\end{tabular*}
\tabnote{Best results are in \textbf{bold}; second-best results are \underline{underlined}.}
\end{table*}

\begin{table*}[!t]
\centering
\caption{Comparison on DroneVehicle.}
\label{tab:dv}
\tablebody
\begin{tabular*}{\textwidth}{@{\extracolsep{\fill}}llccccccc@{}}
\toprule
Method & Venue & \multicolumn{2}{c}{Overall} & \multicolumn{5}{c@{}}{AP$_{50}$ by category} \\
\cmidrule(lr){3-4}\cmidrule(l){5-9}
 & & mAP$_{50}$ & mAP$_{50\text{-}95}$ & Car & Truck & Bus & Van & Freight \\
\midrule
TSFADet~\cite{ref_tsfadet} & ECCV'22 & 73.1 & 44.1 & 89.9 & 67.9 & 89.8 & 54.0 & 63.7 \\
C$^2$Former~\cite{ref_c2former} & TGRS'24 & 74.2 & 47.3 & 90.2 & 68.3 & 89.8 & 58.5 & 64.4 \\
IV-YOLO~\cite{ref_ivyolo} & Sensors'24 & 74.6 & 56.8 & \textbf{97.2} & 65.4 & \underline{94.3} & 53.0 & 63.1 \\
LCAFNet~\cite{ref_lcafnet} & PR'26 & 76.2 & -- & 90.3 & 69.7 & 89.7 & 63.4 & 67.9 \\
SLBAF-Net~\cite{ref_slbafnet} & MTA'23 & 77.0 & 49.5 & 90.2 & 75.9 & 89.9 & 59.9 & 68.6 \\
DMM~\cite{ref_dmm} & arXiv'24 & 77.2 & 55.8 & 90.4 & 77.8 & 88.7 & 66.0 & 63.0 \\
CSOM-ODAF~\cite{ref_csomodaf} & CVPRW'24 & 77.5 & 52.5 & 92.1 & 75.2 & 89.8 & 61.8 & 68.8 \\
GLFNet~\cite{ref_glfnet} & GRSL'24 & 71.4 & 54.8 & 90.3 & 72.7 & 88.0 & 52.6 & 53.6 \\
Fusion-Mamba~\cite{ref_fusionmamba} & TMM'25 & 79.2 & 56.0 & \underline{96.7} & 80.2 & \textbf{95.3} & 64.0 & 59.8 \\
WaveMamba~\cite{ref_wavemamba} & ICCV'25 & 79.8 & \underline{60.5} & 95.0 & 80.4 & 90.6 & 64.5 & 68.5 \\
SLGNet~\cite{ref_slgnet} & arXiv'26 & \underline{80.7} & 57.2 & 96.1 & \underline{80.9} & 91.8 & \underline{65.3} & \underline{69.4} \\
\midrule
\textbf{\methodname} & -- & \textbf{81.5} & \textbf{61.5} & 96.4 & \textbf{81.6} & 92.3 & \textbf{66.1} & \textbf{71.2} \\
\bottomrule
\end{tabular*}
\tabnote{Best results are in \textbf{bold}; second-best results are \underline{underlined}.}
\end{table*}

\subsection{Datasets and Metrics}
We evaluate \methodname on four public RGB--IR detection benchmarks that collectively span a wide range of illumination conditions, scene types, and target scales.

\noindent\textbf{LLVIP}~\cite{ref_llvip} is a low-light pedestrian detection dataset containing 12,025 training and 3,463 test aligned RGB--IR pairs with one category (Pedestrian), making it a focused testbed for nighttime cross-modal fusion.

\noindent\textbf{M3FD}~\cite{ref_m3fd} covers daytime, nighttime, and foggy conditions with 4,200 image pairs across 6 categories (People, Car, Bus, Motorcycle, Lamp, Truck), posing challenges from diverse weather and illumination variation.

\noindent\textbf{DroneVehicle}~\cite{ref_dronevehicle} provides 56,878 drone-view image pairs with 5 vehicle categories (Car, Truck, Bus, Van, Freight Car), featuring small and densely packed targets that stress localization precision.

\noindent\textbf{FLIR-Aligned}~\cite{ref_flir_adas} contains 5,142 aligned image pairs for autonomous driving with 3 categories (Person, Car, Bicycle), representing a scene-cluttered benchmark with significant intra-class scale variation.

We report mAP$_{50}$ and mAP$_{50\text{-}95}$ following the standard COCO evaluation protocol. AP values are reported in percentage points unless otherwise specified.

The experiments are organized to validate the three stages of the proposed register lifecycle. First, comparisons on four benchmarks evaluate whether register-mediated fusion improves detection accuracy across low-light surveillance, multi-weather driving, drone-view traffic scenes, and aligned autonomous-driving data. Second, component and injection-layer ablations examine whether the Bridge stage benefits from \rwpr, \rcrs, and shallow-to-intermediate register injection. Third, \mac ablations and qualitative visualizations assess whether the register-level cross-modal summary can be effectively projected back to spatial feature calibration for detection.

\subsection{Implementation Details}
We use dual-stream frozen DINOv3-B (Base)~\cite{ref_dinov3} ($\sim$86\,M parameters per stream, not updated) as the backbone, initialized from the same pretrained weights for both modalities. The \rwpr bridge is inserted at layers $\{2,5,8\}$ with single-head cross-attention, while multi-scale features for the detection head are extracted at layers $\{5,8,11\}$. This separation is deliberate: injection layers determine where cross-modal register exchange occurs, whereas feature layers determine which backbone outputs are consumed by the detector. The residual gate $\alpha$ is initialized at 0.1 and the \rcrs balance $\beta$ at 0.5. \mac operates on a three-level feature pyramid (256-d) with zero-initialized affine heads, and the detection head is RT-DETR~\cite{ref_rtdetr}. All experiments are conducted on NVIDIA RTX 4090 GPUs. Unless otherwise noted, parameter counts refer to trainable parameters and exclude the two frozen DINOv3-B streams.

\subsection{Comparison with State-of-the-Art}

\subsubsection{Results on M3FD}
Table~\ref{tab:m3fd} reports results on M3FD with per-category breakdown. M3FD is a challenging multi-weather benchmark where illumination and weather conditions vary significantly across scenes, making it a demanding testbed for cross-modal robustness. \methodname achieves the best overall mAP$_{50}$ (92.6\%) and mAP$_{50\text{-}95}$ (64.9\%), surpassing WaveMamba by 0.5 points on both metrics. The per-category results show broad gains on People, Car, Bus, and Motorcycle, while remaining competitive on Lamp and Truck, indicating that the proposed bridge improves cross-modal robustness without concentrating gains in a single category.

\subsubsection{Results on DroneVehicle}
Table~\ref{tab:dv} shows results on DroneVehicle with per-category breakdown. This dataset features small, densely packed vehicles from a drone perspective, posing simultaneous challenges for localization precision and cross-modal alignment. \methodname achieves 81.5\% mAP$_{50}$ and 61.5\% mAP$_{50\text{-}95}$, surpassing the strongest compared baselines by 0.8 points on mAP$_{50}$ and 1.0 point on mAP$_{50\text{-}95}$. Category-level gains are most visible on Truck, Van, and Freight Car, categories that are thermally ambiguous and benefit most from register-level cross-modal context.

\subsubsection{Results on LLVIP}
Table~\ref{tab:llvip} compares \methodname with recent methods on LLVIP. Our method achieves the highest mAP$_{50\text{-}95}$ of 70.5\%, surpassing the strongest compared baseline GM-DETR by 0.3 points, while also leading in mAP$_{50}$. Notably, \methodname uses 27.8\,M trainable parameters, substantially fewer than dense-fusion methods such as Fusion-Mamba (287.6\,M) and CFT (206.0\,M), while delivering higher accuracy than smaller adapter-style baselines such as LCAFNet (15.4\,M) and SLGNet (12.1\,M). This accuracy-efficiency trade-off shows that register-mediated fusion achieves strong low-light detection without the parameter overhead of dense patch-level interaction.

\begin{table}[!t]
\centering
\caption{Comparison on LLVIP.}
\label{tab:llvip}
\tablebody
\begin{tabular*}{\columnwidth}{@{\extracolsep{\fill}}lcccc@{}}
\toprule
Method & Venue & Params & mAP$_{50}$ & mAP$_{50\text{-}95}$ \\
\midrule
ICAFusion~\cite{ref_icafusion} & PR'24 & 108.6M & 95.2 & 60.1 \\
UniRGB-IR~\cite{ref_unirgbir} & MM'25 & 8.9M & 96.1 & 63.2 \\
CFT~\cite{ref_cft} & arXiv'21 & 206.0M & 97.5 & 63.6 \\
LRAF-Net~\cite{ref_lrafnet} & TNNLS'23 & 18.8M & 97.9 & 63.6 \\
Fusion-Mamba~\cite{ref_fusionmamba} & TMM'25 & 287.6M & 97.0 & 64.3 \\
LCAFNet~\cite{ref_lcafnet} & PR'26 & 15.4M & 97.7 & 65.0 \\
COFNet~\cite{ref_cofnet} & TMM'25 & 90.2M & 97.7 & 65.9 \\
WaveMamba~\cite{ref_wavemamba} & ICCV'25 & 69.1M & 98.3 & 66.0 \\
SLGNet~\cite{ref_slgnet} & arXiv'26 & 12.1M & \underline{98.3} & 66.1 \\
DAMSDet~\cite{ref_damsdet} & ECCV'24 & -- & 97.9 & 69.6 \\
GM-DETR~\cite{ref_gmdetr} & CVPRW'24 & 70M & 97.4 & \underline{70.2} \\
\midrule
\textbf{\methodname} & -- & \textbf{27.8M} & \textbf{98.4} & \textbf{70.5} \\
\bottomrule
\end{tabular*}
\tabnote{Trainable params; best/second-best: \textbf{bold}/\underline{underlined}.}
\end{table}

\subsubsection{Results on FLIR-Aligned}
Table~\ref{tab:flir} presents results on FLIR-Aligned, a challenging autonomous-driving dataset with cluttered backgrounds and significant intra-class scale variations. \methodname achieves 88.9\% mAP$_{50}$ and 49.8\% mAP$_{50\text{-}95}$, surpassing the strongest compared baselines by 0.5 points on both metrics. Compared to methods with similar parameter budgets, including SLGNet (12.1\,M, 85.8/45.1) and LCAFNet (15.4\,M, 81.3/41.1), \methodname (27.8\,M) delivers substantially higher accuracy, indicating that the additional parameters in the register bridge yield measurable returns in cluttered autonomous-driving scenes.

\begin{table}[!t]
\centering
\caption{Comparison on FLIR-Aligned.}
\label{tab:flir}
\tablebody
\begin{tabular*}{\columnwidth}{@{\extracolsep{\fill}}lcccc@{}}
\toprule
Method & Venue & Params & mAP$_{50}$ & mAP$_{50\text{-}95}$ \\
\midrule
CFT~\cite{ref_cft} & arXiv'21 & 206.0M & 78.7 & 40.2 \\
ICAFusion~\cite{ref_icafusion} & PR'24 & 108.6M & 79.2 & 41.4 \\
CrossFormer~\cite{ref_crossformer} & PRL'24 & 340.0M & 79.3 & 42.1 \\
LRAF-Net~\cite{ref_lrafnet} & TNNLS'23 & 18.8M & 80.5 & 42.8 \\
LCAFNet~\cite{ref_lcafnet} & PR'26 & 15.4M & 81.3 & 41.1 \\
UniRGB-IR~\cite{ref_unirgbir} & MM'25 & 8.9M & 81.4 & 44.1 \\
COFNet~\cite{ref_cofnet} & TMM'25 & 90.2M & 83.6 & 44.6 \\
GM-DETR~\cite{ref_gmdetr} & CVPRW'24 & 70M & 83.9 & 45.8 \\
SLGNet~\cite{ref_slgnet} & arXiv'26 & 12.1M & 85.8 & 45.1 \\
Fusion-Mamba~\cite{ref_fusionmamba} & TMM'25 & 287.6M & 84.9 & 47.0 \\
DAMSDet~\cite{ref_damsdet} & ECCV'24 & -- & 86.6 & \underline{49.3} \\
WaveMamba~\cite{ref_wavemamba} & ICCV'25 & 69.1M & \underline{88.4} & 48.1 \\
\midrule
\textbf{\methodname} & -- & \textbf{27.8M} & \textbf{88.9} & \textbf{49.8} \\
\bottomrule
\end{tabular*}
\tabnote{Trainable params; best/second-best: \textbf{bold}/\underline{underlined}.}
\end{table}

\subsubsection{Cross-Benchmark Observations}
Across the four benchmarks, the gains of \methodname are consistent but arise from different data characteristics. On LLVIP, the improvement mainly reflects stronger low-light pedestrian localization, where infrared cues compensate for weak visible contrast. On M3FD, the gains are spread across weather and object categories, suggesting that the register bridge does not overfit to a single modality-dominant condition. On DroneVehicle, the improvement on Truck, Van, and Freight Car indicates that register-mediated global context is useful when small aerial targets are visually compact and locally ambiguous. On FLIR-Aligned, the gain under cluttered driving scenes shows that the proposed calibration can suppress background responses while preserving target-relevant thermal and visible cues. These complementary trends indicate that the proposed register pathway improves robustness across modality imbalance, scale variation, and background clutter rather than specializing to a single benchmark.

A notable pattern is that \methodname improves both mAP$_{50}$ and mAP$_{50\text{-}95}$ on all four datasets. The latter metric is more sensitive to localization quality, so the improvement cannot be explained only by stronger classification confidence.

\subsubsection{Efficiency of the Register Pathway}
The accuracy gains above are obtained with a lightweight register pathway rather than dense patch-wise cross-attention. With $k{=}4$ register tokens per stream, \rwpr uses $6kN$ pairwise register-to-patch attention terms across three bidirectional injection layers, which constitutes a small fraction of the $N^2$ terms required by a single dense patch-to-patch cross-attention layer, providing a structural efficiency argument for register-mediated cross-modal communication.

\subsection{Ablation Studies}
All ablations are conducted on LLVIP with the DINOv3-B backbone under the same training schedule unless otherwise noted. Each study isolates one design dimension of the register lifecycle while holding the remaining components fixed.

\subsubsection{Register Lifecycle Component Ablation}
Table~\ref{tab:ablation_component} shows the incremental contribution of each module. The baseline uses frozen dual-stream DINOv3-B with simple concatenation fusion. Adding \rwpr brings +2.0 mAP$_{50\text{-}95}$ by enabling cross-modal information exchange through registers; \rcrs further improves by +1.2 points without adding parameters, confirming that explicit consensus-residual decomposition refines the quality of exchanged information; \mac contributes another +2.1 points by translating the register summary into spatially adaptive patch calibration. The full lifecycle yields a cumulative gain of +5.3 points while keeping the trainable budget at 27.8\,M parameters, confirming the parameter efficiency of the proposed design.

\begin{table}[!t]
\centering
\caption{Component ablation on LLVIP.}
\label{tab:ablation_component}
\tablebody
\begin{tabular*}{\columnwidth}{@{\extracolsep{\fill}}ccccccc@{}}
\toprule
\rwpr & \rcrs & \mac & Params & mAP$_{50}$ & mAP$_{50\text{-}95}$ & $\Delta$ \\
\midrule
 &  &  & 11.8M & 97.1 & 65.2 & -- \\
Patch-Query &  &  & 26.0M & 97.3 & 65.8 & +0.6 \\
\checkmark &  &  & 26.0M & 97.6 & 67.2 & +2.0 \\
\checkmark & \checkmark &  & 26.0M & 97.8 & 68.4 & +3.2 \\
\checkmark & \checkmark & \checkmark & 27.8M & \textbf{98.4} & \textbf{70.5} & +5.3 \\
\bottomrule
\end{tabular*}
\tabnote{Params reports trainable parameters added on top of the two frozen DINOv3-B backbones (172.0\,M, not updated). \rcrs adds only 3 scalar parameters and does not change the rounded trainable count. Patch-Query replaces register tokens with $k{=}4$ uniformly sampled patch tokens as cross-attention queries, keeping all other settings identical.}
\end{table}

\begin{figure}[!t]
\centering
\includegraphics[width=0.82\columnwidth]{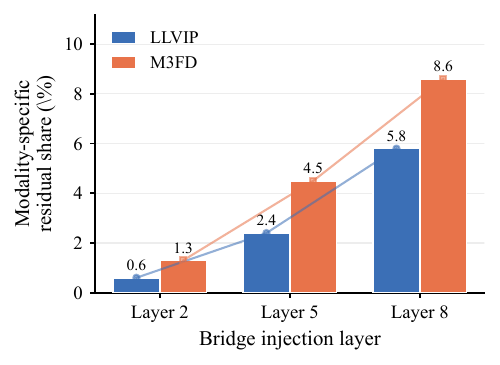}
\caption{Modality-specific residual energy share of bridged registers on LLVIP and M3FD. Registers remain dominated by consensus ($>90\%$), while the residual share exposed by \rcrs grows with depth.}
\label{fig:rcrs_energy}
\end{figure}

\subsubsection{Backbone Freezing and Injection Layer Analysis}
Table~\ref{tab:ablation_design} (top) compares frozen and unfrozen backbone variants, and also tests whether the bridge transfers across DINOv3 pretraining versions. \methodname with a frozen DINOv3-B backbone achieves 70.5 mAP$_{50\text{-}95}$, slightly exceeding the fully fine-tuned variant (70.3), confirming that the pretrained register prior is worth preserving rather than overwriting through fine-tuning. Replacing DINOv3-B with DINOv2-B yields 69.8 mAP$_{50\text{-}95}$, a drop of only 0.7 points despite the weaker pretraining, indicating that the register bridge generalizes across backbone versions rather than relying on DINOv3-specific representations. The table (bottom) examines bridge placement: a single bridge at layer~8 already provides +2.3 over the concat baseline, and progressively adding shallower bridges reaches 70.5 with $\{2,5,8\}$. Adding a bridge at layer~11 degrades performance by 0.7, because registers modified at the final layer have no subsequent self-attention through which to propagate their updates back into the patch representation.

\begin{table}[!t]
\centering
\caption{Backbone freezing and pretraining version (top) and injection-layer placement (bottom) on LLVIP.}
\label{tab:ablation_design}
\tablebody
\begin{tabular*}{\columnwidth}{@{\extracolsep{\fill}}lccc@{}}
\toprule
Setting & Params & mAP$_{50}$ & mAP$_{50\text{-}95}$ \\
\midrule
Unfrozen, concat            & 183\,M & 97.8 & 69.4 \\
Unfrozen, \methodname       & 199\,M & 98.3 & 70.3 \\
Frozen, \methodname (DINOv2-B) & 28\,M & 98.1 & 69.8 \\
Frozen, \methodname (DINOv3-B) & 28\,M & \textbf{98.4} & \textbf{70.5} \\
\midrule
\multicolumn{4}{@{}l}{\small\textit{Injection layers (multi-scale feature layers fixed at $\{5,8,11\}$)}} \\
\midrule
$\{8\}$             & -- & 97.4 & 67.5 \\
$\{5, 8\}$          & -- & 97.8 & 69.0 \\
$\{2, 5, 8\}$       & -- & \textbf{98.4} & \textbf{70.5} \\
$\{2, 5, 8, 11\}$   & -- & 98.0 & 69.8 \\
\bottomrule
\end{tabular*}
\end{table}

\subsubsection{RCRS and MAC Design Choices}
Table~\ref{tab:ablation_modules} (top) studies the consensus--residual split in \rcrs. Forcing full consensus ($\beta{=}0$) discards modality-specific residuals and degrades mAP$_{50\text{-}95}$ by 0.4 points; making $\beta$ learnable recovers and surpasses the full-consensus variant. The net gain of \rcrs over \rwpr alone is +1.2 points (Table~\ref{tab:ablation_component}), indicating that the model benefits from retaining a calibrated amount of modality-specific information. Fig.~\ref{fig:rcrs_energy} further shows that the residual share grows with depth, confirming that the cues \rcrs separates become increasingly informative in deeper layers. The table (bottom) compares calibration strategies in \mac: channel-only (FiLM-style) modulation improves by +1.2 over no calibration, and replacing it with spatially adaptive calibration adds a further +0.9, demonstrating that location-specific modulation is more effective than global channel scaling.

\begin{table}[!t]
\centering
\caption{\rcrs consensus--residual design (top) and \mac calibration design (bottom) on LLVIP.}
\label{tab:ablation_modules}
\tablebody
\begin{tabular*}{\columnwidth}{@{\extracolsep{\fill}}lcc@{}}
\toprule
Setting & mAP$_{50}$ & mAP$_{50\text{-}95}$ \\
\midrule
No \rcrs (\rwpr only)      & 97.6 & 67.2 \\
\rcrs, full consensus      & 97.5 & 66.8 \\
\rcrs, learnable $\beta$   & \textbf{98.4} & \textbf{70.5} \\
\midrule
\multicolumn{3}{@{}l}{\small\textit{\mac calibration strategy (starting from \rwpr+\rcrs)}} \\
\midrule
No \mac                         & 97.8 & 68.4 \\
\mac channel-only (FiLM-style)  & 98.0 & 69.6 \\
\mac spatially adaptive         & \textbf{98.4} & \textbf{70.5} \\
\bottomrule
\end{tabular*}
\end{table}

\subsection{Qualitative Analysis}

\subsubsection{Detection Results}
\begin{figure*}[!t]
\centering
\includegraphics[width=0.98\textwidth]{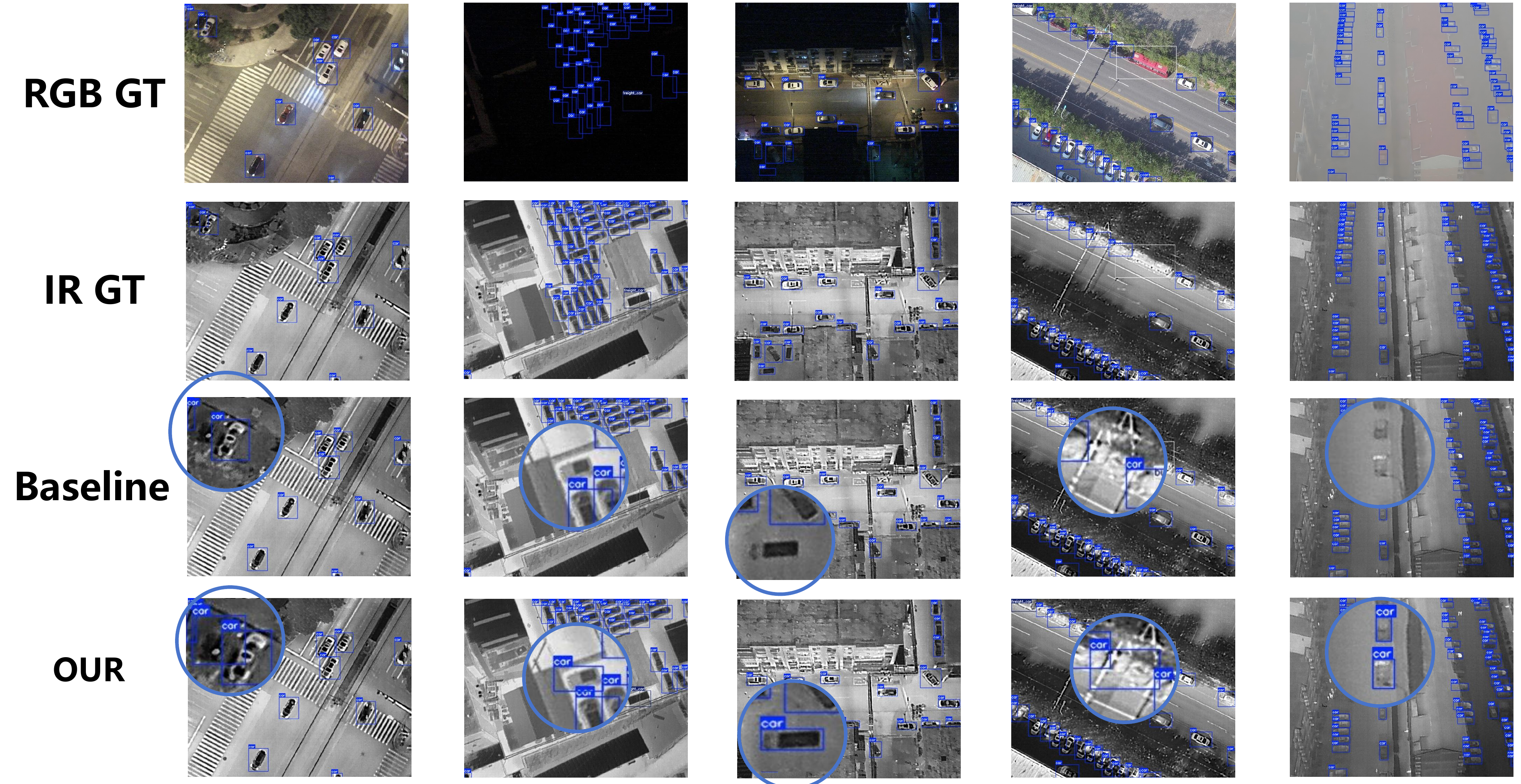}
\caption{Qualitative detection results on representative RGB--IR pairs. Rows show RGB input, IR input, the frozen simple-concatenation baseline, and \methodname. Compared with the baseline, \methodname recovers missed targets and suppresses false positives, especially for small and occluded objects.}
\label{fig:det_vis}
\end{figure*}

Fig.~\ref{fig:det_vis} shows qualitative detection results on four representative RGB--IR pairs spanning diverse illumination and scene conditions. Compared to the frozen baseline, \methodname produces fewer missed detections and false positives, with the most pronounced improvements on occluded and small targets where cross-modal complementarity is most critical.

\subsubsection{Feature Attention Visualization}
\begin{figure}[!t]
\centering
\includegraphics[width=\columnwidth]{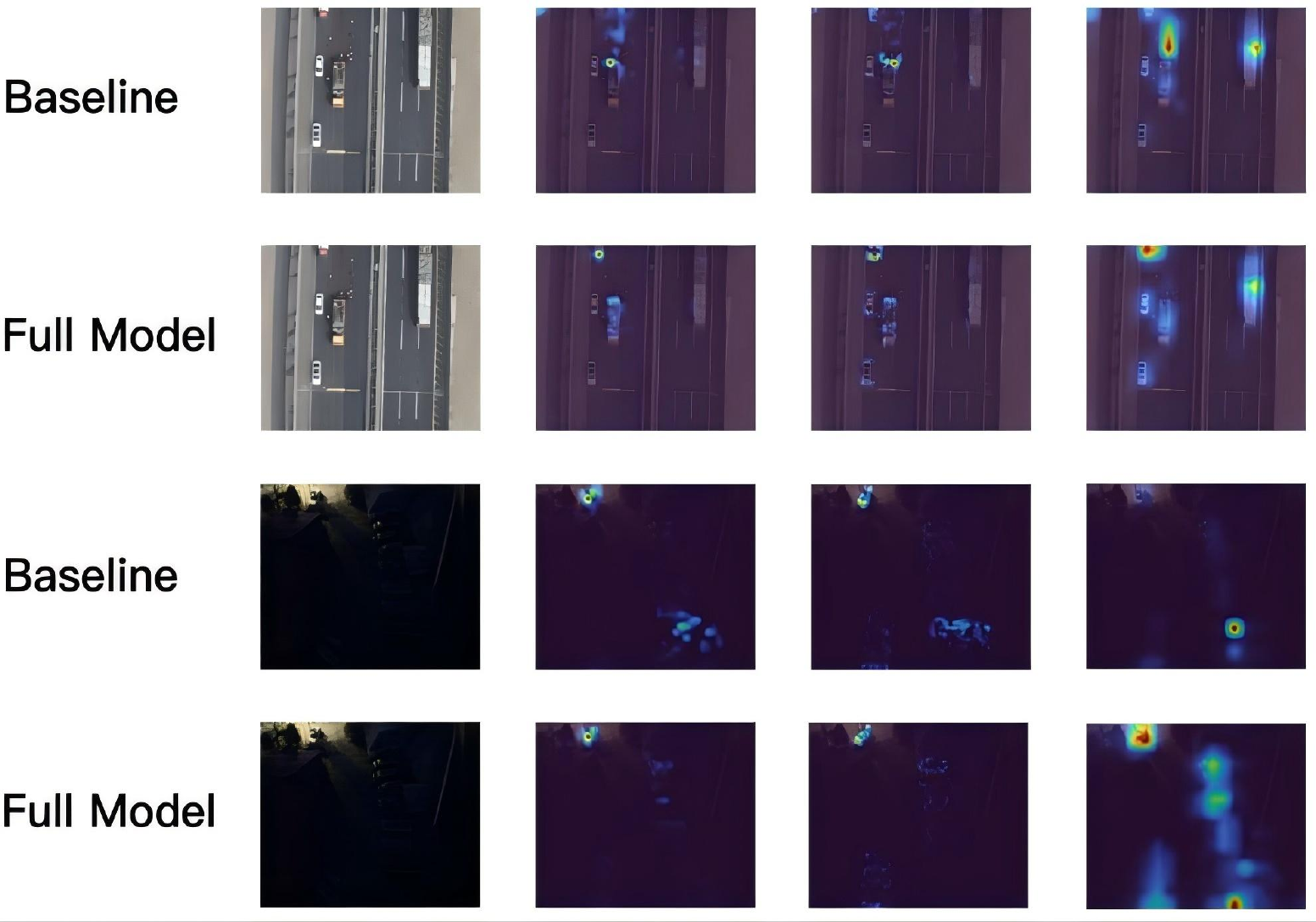}
\caption{Attention comparison between the frozen baseline and \methodname at the multi-scale feature layers. Rows show two scenes; columns show the input image and attention maps at layers~5, 8, and 11. \methodname concentrates responses on target-relevant regions more selectively than the baseline.}
\label{fig:attn_vis}
\end{figure}

Fig.~\ref{fig:attn_vis} compares the attention maps of the frozen baseline and \methodname at the three multi-scale feature layers $\{5,8,11\}$ consumed by the detection head. \methodname attends more coherently to target-relevant regions: responses are better aligned with object extents and suppress background clutter, whereas the baseline distributes attention more diffusely across non-target areas. The improvement is consistent across all three depths, suggesting that register-mediated fusion yields cleaner multi-scale features rather than benefiting only a single scale.

%% file: sections/conclusion.tex
\section{Conclusion}
\label{sec:conclusion}
RegisterBridgeMM demonstrates that pretrained DINOv3 register tokens carry a latent modality-shared/modality-specific structure suitable for cross-modal communication. Routing RGB--IR interaction through this register bottleneck keeps both backbones frozen while achieving strong detection performance across four benchmarks.